%% file: acl_latex.tex
\pdfoutput=1

\documentclass[11pt]{article}

\usepackage{acl}

\usepackage{times}
\usepackage{latexsym}
\usepackage{dirtytalk}
\usepackage{amsmath}
\usepackage{csquotes}
\usepackage{graphicx}
\usepackage{multirow}
\usepackage{makecell}
\usepackage{booktabs}
\usepackage{enumitem}
\usepackage{hyperref}
\usepackage{pgfplots}
\usepackage{tabularx}
\usepackage{longtable}

\usepackage[T1]{fontenc}

\usepackage[utf8]{inputenc}

\usepackage{microtype}

\DeclareMathAlphabet\mathbfcal{OMS}{cmsy}{b}{n}
\makeatletter
\def\@fnsymbol#1{\ensuremath{\ifcase#1\or \dagger\or *\or \ddagger\or
   \mathsection\or \mathparagraph\or \|\or **\or \dagger\dagger
   \or \ddagger\ddagger \else\@ctrerr\fi}}

\title{Meet Your Favorite Character: Open-domain Chatbot Mimicking \\Fictional Characters with only a Few Utterances}

\author{Seungju Han\thanks{\; Equal contribution}\;\;\;\; Beomsu Kim\footnotemark[1]\;\;\;\; Jin Yong Yoo\footnotemark[1]\;\;\;\; Seokjun Seo \\ \textbf{Sangbum Kim}\;\;\;\; \textbf{Enkhbayar Erdenee}\;\;\;\; \textbf{Buru Chang}\thanks{\; Corresponding author} \\
  Hyperconnect \\
  \{\small\texttt{seungju.han,beomsu.kim,jeffrey,seokjun.seo,airdish,enkhbayar.erdenee,buru.chang\}@hpcnt.com} \\
}

\begin{document}
\maketitle
\input{Sections/0_Abstract}
\input{Sections/1_Introduction}
\input{Sections/2_Method}
\input{Sections/3_Experiments}
\input{Sections/4_Results}
\input{Sections/5_Conclusion}
\input{Sections/6_EthicalConsiderations}

\bibliography{anthology,custom}
\bibliographystyle{acl_natbib}

\newpage
\appendix
\input{Sections/Appendix}

\end{document}

%% file: Sections/0_Abstract.tex
\begin{abstract}\label{sec:0_abstract}
In this paper, we consider mimicking fictional characters as a promising direction for building engaging conversation models.
To this end, we present a new practical task where only a few utterances of each fictional character are available to generate responses mimicking them.
Furthermore, we propose a new method named Pseudo Dialog Prompting (PDP) that generates responses by leveraging the power of large-scale language models with prompts containing the target character's utterances.
To better reflect the style of the character, PDP builds the prompts in the form of dialog that includes the character's utterances as dialog history.
Since only utterances of the characters are available in the proposed task, PDP matches each utterance with an appropriate pseudo-context from a predefined set of context candidates using a retrieval model.
Through human and automatic evaluation, we show that PDP generates responses that better reflect the style of fictional characters than baseline methods.
\end{abstract}

%% file: Sections/1_Introduction.tex
\section{Introduction}\label{sec:1_introduction}
\textit{How would you feel if you could talk to your favorite character?}

In recent years, open-domain conversation models~\cite{adiwardana2020towards,roller2021recipes} have achieved remarkable progress with the development of large-scale language models~\cite{radford2019language,brown2020language}.
Meanwhile, recent studies have suggested several directions reflecting desirable traits of real-life conversation to make open-domain conversation models more engaging beyond plain chit-chat.
Style-controlling conversation models generate responses in the target styles such as emotion~\cite{zhou2018emotional,demszky2020goemotions} and empathy~\cite{rashkin2019towards}.
Persona-grounded conversation models~\cite{zhang2018personalizing,kim2020will,majumder2020like} produce responses that preserve consistent personalities by leveraging personal descriptions (e.g., "I have two dogs").
In this paper, we consider \textit{mimicking fictional characters} as a promising direction for building engaging conversation models.

\input{Figures/Figure_1}
When it comes to building conversation models that mimic fictional characters, two major challenges prevent us from directly applying previous models designed for conditional response generation:
(1) It is \textit{difficult to define fictional characters} with only a few descriptions, as in persona-grounded conversation models.
Furthermore, it is not expressive enough to represent characters' styles with discrete labels (e.g., angry, happy), as style-controlling conversation models do.
(2) There \textit{lacks sufficient dialog data of fictional characters} for training conversation models.
It is inefficient to manually create dialog datasets of characters for training, especially considering that additional data is needed for each new character.

To address these two challenges, we propose a new task where only a few utterances of the fictional characters are available to generate responses mimicking the characters. 
Such setting is justified by the two following reasons: 
(1) Utterances of fictional characters provide useful clues for generating responses mimicking the characters as the personal traits or styles of speakers are inherent in their utterances~\cite{boyd2020large,li2020aloha}. 
(2) Collecting only a few utterances of target characters is a cost-effective scenario compared to constructing the full dialog data consisting of context and utterance pairs; this allows us to extend our method to a new character easily.

To perform the task, we introduce \textbf{Pseudo Dialog Prompting} (PDP), a method that builds prompts using a few numbers of target characters' utterances to leverage the power of pre-trained language models.
We claim that designing the prompt in the form of dialog that includes the character's utterances as dialog history (as in Figure~\ref{fig:1}) is an effective method for reflecting the style of character.
However, since only utterances of the characters are available in the proposed task, we match each utterance with an appropriate pseudo-context by using a retrieval model~\cite{humeau2019poly} to select the relevant context from a predefined set of context candidates.
Through human and automatic evaluation, we show that PDP generates responses that better reflect the style of fictional characters than existing baseline models.

%% file: Figures/Figure_1.tex
\begin{figure}[t]
\centering
\includegraphics[width=0.95\columnwidth]{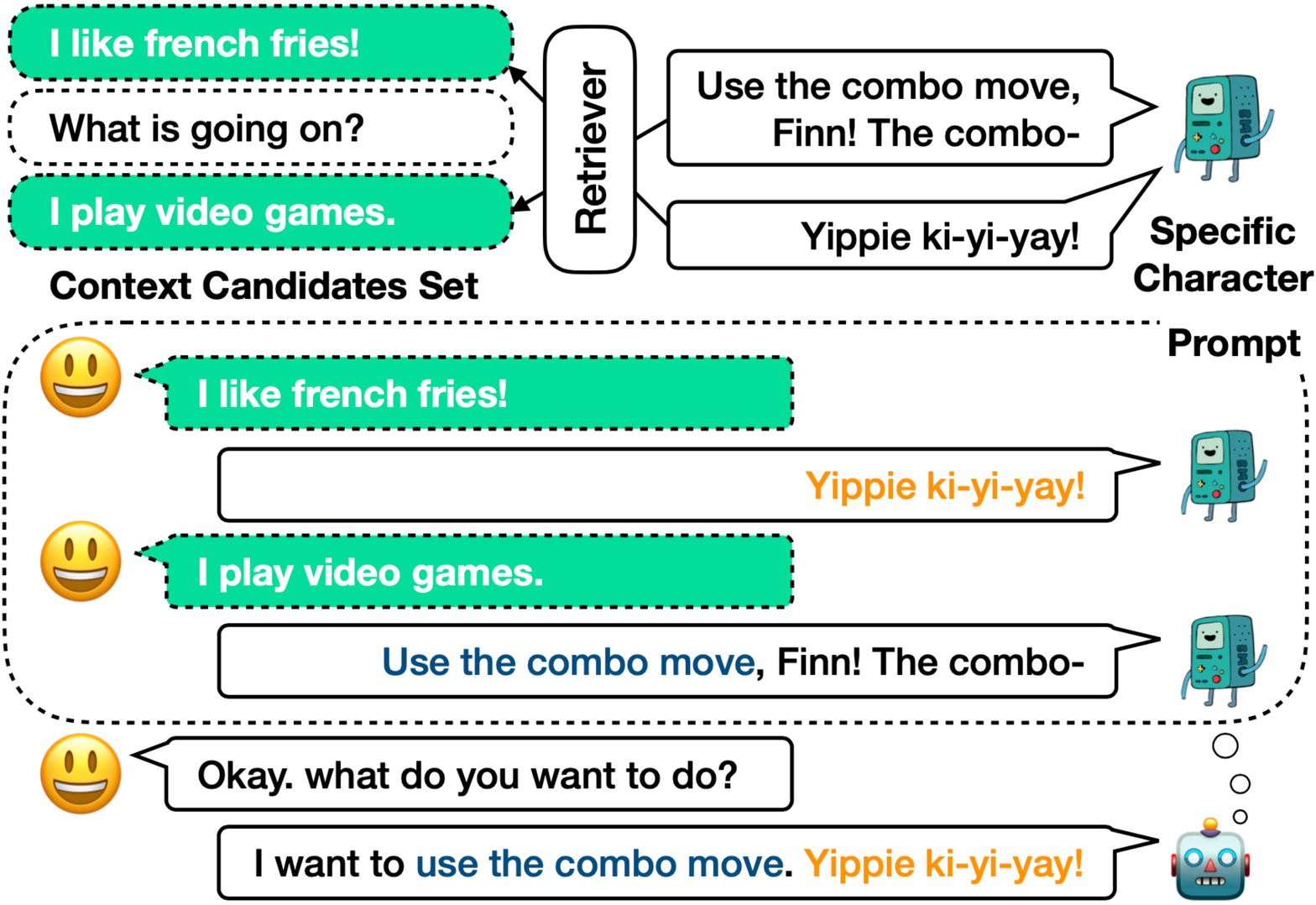}
\caption{Illustration of PDP. The retriever matches pseudo-context for utterances from the character, and utilizes them in a prompt while generating the response.}
\vspace*{-4mm}
\label{fig:1}
\end{figure}

%% file: Sections/2_Method.tex
\section{Method}\label{sec:2_method}
We model a conversation agent that generates a response $r$ corresponding to a given context $x$ while mimicking an arbitrary character with $k$ utterances $\{u_1, u_2, \cdots, u_k\}$ of the character.
The simplest way to design the prompt with the character's utterances is to concatenate utterances as \citet{madotto2021few} does for PersonaChat~\cite{zhang2018personalizing}.
However, in our preliminary experiments, we observed that this method tends to generate dull responses that do not reflect the styles of the character (will be shown in Section~\ref{sec:4_results}).
We hypothesize that the language model fails to utilize the utterances because such a format of the prompt is unlikely to have appeared naturally in the training set~\citep{brown2020language,wei2021finetuned}.

To address this issue, we propose PDP, which builds a dialog format prompt where character utterances are included in the dialog history, as depicted in Figure 1.
Since a speaker tends to maintain a consistent style throughout the conversation, using such a prompt will induce the language model to generate responses that seamlessly reflect the style from the character's utterances.
To build a dialog when only given the utterances of the character, we require a pseudo-context $c_i$ matching each utterance $u_i$ to get a context-utterance pair $(c_i, u_i)$.
We use a retriever $R$ to acquire a pseudo-context $c_i$.
Particularly, we employ Bi-encoder~\cite{humeau2019poly} as our retriever $R$.
We first define a fixed set of single-turn context candidates $\mathcal{C}$ obtained from BST dataset~\cite{smith2020can}, which is the largest open-domain conversation dataset released to date.
We then select a candidate as the pseudo-context $c_i$ for the given utterance $u_i$ using $R$.
Bi-encoder maps the context $c$ and the response $r$ into the embedding space as $e_{\text{ctx}}(c)$ and $e_{\text{resp}}(r)$, respectively.
Bi-encoder is trained to represent the relevance score between a context $c$ and response $r$ with $e_{\text{ctx}}(c) \cdot e_{\text{resp}}(r)$.
There are several variants to select the pseudo-context $c_i$ as follows:

\begin{itemize}[leftmargin=*, topsep=0pt]
\item\textbf{Static Match} selects a pseudo-context $c_i$ that can coherently precede the given utterance $u_i$ using the retrieval model $R$.
Given $u_i$, $R$ calculates a score $s_{\text{stat}}$ for each $c \in \mathcal{C}$ by $s_{\text{stat}}(c; u_i) = e_{\text{ctx}}(c) \cdot e_{\text{resp}}(u_i)$.
We set the pseudo-context $c_i$ of $u_i$ as $c_i = \text{argmax}_{c} s_{\text{stat}}(c; u_i)$.
We name this variant \textit{static} since the selected pseudo-context $c_i$ depends only on the given utterance $u_i$.
\vspace{-0.5em}
\item \textbf{Dynamic Match} selects a pseudo-context $c_i$ relevant to the input context $x$ in addition to $u_i$. 
Given $x$ and $u_i$, $R$ calculates a score $s_{\text{dyn}}$ for each $c \in \mathcal{C}$ by $s_{\text{dyn}}(c; x, u_i) = e_{\text{ctx}}(c) \cdot e_{\text{ctx}}(x) + s_{\text{stat}}(c; u_i)$. 
We set the pseudo-context $c_i$ of $u_i$ as $c_i = \text{argmax}_{c} s_{\text{dyn}}(c; x, u_i)$.
Since language models quickly adapt to the context-response mapping of the given prompt via in-context learning, we believe providing pseudo-contexts that are semantically similar to the input context as in Dynamic Match facilitates the reflection of styles in corresponding utterances.
We name this variant \textit{dynamic} because the pseudo-context $c_i$ depends on the varying input context $x$.
\vspace{-0.5em}
\item \textbf{Random Match} selects a pseudo-context $c_i$ randomly from the context candidates set $\mathcal{C}$ without using $R$.
This variant is used as a baseline to study the effect of the pseudo-context $c_i$.
\end{itemize}

Finally, all the $k$ pairs $(c_i, u_i)$ of the character are sorted by $e_{\text{ctx}}(x) \cdot e_{\text{resp}}(u_i)$ in ascending order and are concatenated into a prompt in a dialog format.

%% file: Sections/3_Experiments.tex
\section{Experiments}\label{sec:3_experiments}

\subsection{Evaluation}\label{subsec:dataset}
We employ the \textbf{HLA-Chat}~\cite{li2020aloha} dataset to define the set of characters for evaluation.
HLA-Chat consists of single-turn dialogs of characters in various TV shows.
We select ten characters among all the characters and manually curate eight utterances that best reveal each character's unique characteristics from their utterances in the dataset.
Note that we consider that creating eight utterances is feasible even if new characters are given and we also empirically observed that language models adequately reflect each character's unique characteristics from the eight utterances.

In evaluating the performance of each method, we focus on two criteria: (1) Does the model's response reflect the style of a given character? (2) Does the model respond coherently to the given dialog context?
To examine these two criteria, we run the model on fixed dialog contexts and calculate metrics that exhibit the style reflection and dialog coherency.
We use the utterances of the test split of DailyDialog~\cite{li2017dailydialog} for dialog contexts.
\newline
\textbf{Human Evaluation.}
We conduct a human evaluation to assess the quality of the generated responses.
First, we select five characters which style can be distinguished apparently.
We then randomly sample 50 contexts from the full fixed-context set of the characters. 
Using Amazon MTurk, we collect human annotations for the samples contexts.
Human evaluators are asked to rate from 0 to 2 scale score how each model response (1) strongly reveals the style of a given character (\textit{Style Strength}) and (2) whether a response is fluent and appropriate for a given dialog context (\textit{Appropriateness}).
To reduce annotator bias and inter-annotator variability, we apply Bayesian Calibration~\cite{kulikov2019importance} to the human evaluation score.
\newline
\textbf{Automatic Evaluation.}
Similar to the previous works on text style transfer~\citep{li2018delete, riley2021textsettr, smith2020controlling}, we utilize a character classifier trained on the utterances in HLA-Chat to measure the style strength of the generated responses.
We denote \textit{StyleProb} as the classifier's average probability of predicting a target character.
We use StyleProb instead of Style Accuracy since HLA-Chat has a class imbalance issue so that the performance on infrequent classes are hard to be measured by accuracy.
For measuring coherency, we use \textit{MaUdE}~\citep{sinha2020learning}, an automated dialog evaluation metric known to capture human judgment on the coherency of response.

\subsection{Pre-trained Language Model}
For all the methods, we use a decoder-only transformer of 3.8B parameters, denoted as \textit{Base-LM}, as a base language model.
To make Base-LM acquire general language skills and better understand conversations, we train Base-LM on The Pile~\citep{gao2020pile} and an additional corpus of public web documents.

\subsection{Baseline Methods}\label{subsec:baseline}
\textbf{Only Utterances.}
Instead of utilizing pseudo-context as suggested in our methods, we provide the set of character utterances as the "quotes of character during conversation" in the prompt. 
Comparing PDP with this method will verify the effect of pseudo-contexts.
\newline
\textbf{Zero-shot Prompting.}
In this method, we only include the name of the character and the show in the prompt without using utterances of the character.
The format of the prompt is similar to the prompt of \citet{madotto2021few} for controlled generation.
\newline
\textbf{TextSETTR~\citep{riley2021textsettr}.}
We first construct a dialog prompt similar to Zero-shot Prompting (but without character information) and use it with Base-LM to generate plain responses.
Then, we use TextSETTR, a few-shot text style transfer model that can transfer arbitrary styles without additional training, to transfer the style of plain responses to the target character's style.
\newline
\textbf{GCC~\citep{boyd2020large}.}
GCC is a method to control a user persona by utilizing the user's conversation history by concatenating users' previous utterances before input dialog context.
Still, it has the drawback that it requires further training on a large-size character-conditioned dialog dataset.

\subsection{Advantaged Methods}
Unlike baseline methods that only have access to a few utterances of characters, advantaged methods also have access to additional data, which gives them an unfair advantage over other methods.
\newline
\textbf{HLA-Chat Full-dataset Fine-tuning.}
We fine-tune Base-LM on the full HLA-chat dataset.
In this method, character information is injected by concatenating the character's name and the show's name at the front of the dialog input.
\newline
\textbf{Gold Match.}
Instead of using pseudo-context, this model uses the actual contexts corresponding to character example utterances annotated in the HLA-chat dataset.

Details for all methods and experiments are further described in Appendix.

%% file: Sections/4_Results.tex
\section{Results}\label{sec:4_results}
\input{Tables/1_Main_Results}
\input{Tables/2_Character_Examples}
Table~\ref{tab:main_result} shows the experimental results.
Overall, our proposed PDP demonstrates far better style reflection scores on both human evaluation and automated metrics than all baseline methods -- and even better than advantaged methods.
In particular, PDP shows significantly higher style reflection scores compared to \textit{Only Utterances}.
Considering that the core difference between the prompt of PDP and that of \textit{Only Utterances} is the presence of pseudo-contexts, this result demonstrates that providing a dialog-formatted prompt is highly effective at reflecting the styles of a character.

While PDP methods generally report better style reflection scores than baseline methods, we observe that the performance on style reflection and response coherency varies to some extent depending on how pseudo-context is selected.
\textit{Static Match} shows the highest response coherency scores among all variants of PDP while performing a little bit worse than \textit{Dynamic Match} in terms of style reflection metrics.
On the other hand, \textit{Dynamic Match} shows the best performance on style reflection metrics, where it losses some coherency.
This observation confirms our hypothesis that using pseudo-context $c_i$ that is semantically similar to the input context $x$ is effective for utilizing styles from the character's utterances.
Thus, the choice between \textit{Static Match} and \textit{Dynamic Match} depends on which of the two qualities -- style and coherency -- is more important.
Lastly, \textit{Random Match}, which is considered a simple ablation baseline, also shows reasonably high performance in terms of style reflection metrics.
We plan to analyze the \textit{Random Match} method in a follow-up study since it is unexpected that such a simple baseline shows high performance.

\textbf{Discussion.}
Gold Match shows worse performance in style strength than PDP. 
We believe that the gold context-response pairs in the HLA-Chat are not always the most appropriate pairs for our experiments.
Since the HLA-Chat originated from scripts of TV shows, there might be some additional contexts outside of a single-turn dialogue (e.g., the background of characters, events that happened before the dialogue, audio-visual information, etc.). 
Without understanding the context behind the scripts, even gold context-response pairs might seem irrelevant. 
Therefore, directly using the context-response pairs in HLA-Chat as in Gold Match could adversely affect the quality of subsequent responses in style strength and coherency.

PDP methods tend to have slightly lower response coherency scores compared to other baselines.
Our speculations for this phenomenon are as follows.
Pseudo-dialog pairs $(c_i, u_i)$ created by PDP methods might have some degree of incoherency, and it might incur adverse effects in coherency via in-context learning in the language model.
The fact that the response coherency score of \textit{Static Match} is higher compared to \textit{Dynamic Match}, which finds a pseudo-context that is more similar to the input context, or Random Match, which finds a random pseudo context at all, supports this claim.
Additionally, automated metrics like MaUdE are tuned to work with texts in standard dialog style.
Since responses that strongly reflect character styles (e.g., \textit{"Yippie ki-yi-yay!"} in Figure~\ref{fig:1}) are out-of-domain examples when put next to standard texts, there might be an unavoidable decrease in MaUdE scores.
An interesting future work would be finding a method that does not reduce response coherency while also successfully reflecting the character styles.

\textbf{Applicability of PDP to other language models.}
\input{Tables/6_DifferentLMs}
We further evaluate our method by leveraging different language models instead of Base-LM to verify that our method generally works well on any language model.
We use three pre-trained language models, GPT-J 6B~\cite{gpt-j}, GPT-Neo 2.7B~\cite{gpt-neo}, and GPT2-xl 1.5B~\cite{radford2019language}, which are publicly available.
Similar to our main experiments, we conduct the automatic evaluation with these language models.

The results are shown in Table~\ref{tab:different_lms}.
The overall trend of the results is similar to the results using Base-LM as a pre-trained language model (Table~\ref{tab:main_result}).
This common trend shows that mimicking characters through the PDP method can be generally used not only with Base-LM but also with other pre-trained language models.

%% file: Tables/1_Main_Results.tex
\begin{table*}[t]\setlength{\tabcolsep}{0.4em}
\centering
\footnotesize
\begin{tabular}{clcccccccccc}
\toprule
\multicolumn{1}{c}{\multirow{2}[2]{*}{Method Type}} & \multicolumn{1}{c}{\multirow{2}[2]{*}{Methods}} & \multicolumn{3}{c}{Human Evaluation (Raw)} & \multicolumn{3}{c}{Human Evaluation (Cali.)} & \multicolumn{2}{c}{Automatic Evaluation} \\ \cmidrule(lr){3-5}  \cmidrule(lr){6-8} \cmidrule(lr){9-10}
& & \multicolumn{1}{c}{Style.} & \multicolumn{1}{c}{Appr.}  & \multicolumn{1}{c}{Sum}  & \multicolumn{1}{c}{Style.} & \multicolumn{1}{c}{Appr.}  & \multicolumn{1}{c}{Sum}  & \multicolumn{1}{c}{StyleProb} & \multicolumn{1}{c}{MaUdE} \\
\midrule
\multirow{4}{*}{\textbf{Baselines}}
& Only Utterances& 1.200& 1.263& 2.463& 1.147 & 1.124 & 2.271 & 0.2098& 0.8887\\
& Zero-shot Prompt& 1.172& 1.236& 2.408 & 1.114 & 1.037 & 2.151 &0.1432& 0.8857\\
& TextSETTR& 1.155& 1.317 &2.472 & 1.060 & 1.059 & 2.119 &  0.1045& 0.8435 \\
& GCC& 1.193& 1.256& 2.449& 1.136 & 1.069 & 2.205  & 0.2391& 0.8835\\
\midrule
\multirow{3}{*}{\textbf{Ours}}
& Static Match & 1.277& 1.229& 2.506 & 1.195 & 1.123 & 2.318 & 0.4192& 0.8467\\
& Dynamic Match& 1.331& 1.203& 2.534 & 1.276 & 1.021 & 2.297 &  0.4789& 0.8299\\
& Random Match& 1.297& 1.195& 2.492& 1.199 & 1.035 & 2.234 & 0.4774& 0.8358\\
\midrule
\midrule
\multirow{2}{*}{\textbf{Advantaged}}
& Full-dataset Fine-tuning& 1.260& 1.329& 2.589& 1.164 & 1.130 & 2.295 & 0.2880& 0.8793\\
& Gold Match &1.287& 1.245& 2.532 & 1.157 & 1.097 & 2.254 &  0.4575& 0.8350\\
\bottomrule
\end{tabular}
\caption{Evaluation results of style reflection metrics and response coherency metrics.
Raw and Cali. indicate raw human evaluation results and Bayesian calibrated scores, respectively.
}
\vspace*{-4mm}
\label{tab:main_result}
\end{table*}

%% file: Tables/2_Character_Examples.tex
\begin{table}[]
\centering
\footnotesize
\setlength
\tabcolsep{2.5pt}
\begin{tabular}{l}
\toprule
\multicolumn{1}{l}{\begin{tabular}[l]{l}\textbf{Input}: We'll pick you up at your place at noon. \\Be there or be square!\end{tabular}} \\ 
\midrule
\midrule
 \multicolumn{1}{l}{\begin{tabular}[l]{l}\textbf{Marge Simpson} (from \textit{The Simpsons}): Homer, you are\\
 my rock! I'm counting on you!\end{tabular}} \\

\cmidrule(lr){1-1}
\multicolumn{1}{l}{\begin{tabular}[l]{l}\textbf{Bart Simpson} (from \textit{The Simpsons}): Lis, are you \\
coming with us, too? \end{tabular}} \\

\cmidrule(lr){1-1}
\multicolumn{1}{l}{\begin{tabular}[l]{l}\textbf{Michael} (from \textit{The Office US}): I will be there. \\
I will be there! I will be there, all right? \\
So, just... be ready. Don't be late. You know, I'm sorry.\end{tabular}} \\

\cmidrule(lr){1-1}
\multicolumn{1}{l}{\begin{tabular}[l]{l}\textbf{Dwight} (from \textit{The Office US}): Micheal, are you listening \\ to me? Are you even paying attention?! \end{tabular}} \\

\cmidrule(lr){1-1}
\multicolumn{1}{l}{\begin{tabular}[l]{l}\textbf{Rachel} (from \textit{Friends}): Oh my god, Phoebe, I just-\end{tabular}} \\


\cmidrule(lr){1-1}
\multicolumn{1}{l}{\begin{tabular}[l]{l}\textbf{Spock} (from \textit{Star Trek}): Aye, Mister Scott. I'll be there.\end{tabular}} \\
\bottomrule
\end{tabular}
\caption{Responses (Other rows) generated from given input (Top row) by \textit{Dynamic Match} for each character.}
\vspace*{-4mm}
\label{tab:2_character_example}
\end{table}

%% file: Tables/6_DifferentLMs.tex
\begin{table}[t]
\centering
\resizebox{0.95\columnwidth}{!}{%
\begin{tabular}{clccc}
\toprule
\multicolumn{1}{c}{\multirow{1}{*}{Pre-trained LM}} & \multicolumn{1}{c}{\multirow{1}{*}{Method}} & \multicolumn{1}{c}{StyleProb} & \multicolumn{1}{c}{MaUdE} \\
\midrule
\multirow{5}[5]{*}{\textbf{GPT-J (6B)}}
& Only Utterances & 0.2200 & 0.8827\\
\cmidrule(lr){2-4}
& Static Match & 0.3805 & 0.8638\\
& Dynamic Match & 0.4166 & 0.8535\\
& Random Match & 0.4045 & 0.8589\\
\cmidrule(lr){2-4}
& Gold Match & 0.3860 & 0.8671\\
\midrule
\multirow{5}[5]{*}{\textbf{GPT-Neo (2.7B)}}
& Only Utterances & 0.1834 & 0.8901\\
\cmidrule(lr){2-4}
& Static Match & 0.3561 & 0.8691\\
& Dynamic Match & 0.3940 & 0.8604\\
& Random Match & 0.3950 & 0.8683\\
\cmidrule(lr){2-4}
& Gold Match & 0.3872 & 0.8732\\
\midrule
\multirow{5}[5]{*}{\textbf{GPT2-xl (1.5B)}}
& Only Utterances & 0.1831 & 0.8817\\
\cmidrule(lr){2-4}
& Static Match & 0.3388 & 0.8736\\
& Dynamic Match  & 0.3760 & 0.8728\\
& Random Match & 0.3515 & 0.8780\\
\cmidrule(lr){2-4}
& Gold Match & 0.3579 & 0.8754\\
\bottomrule
\end{tabular}%
}
\caption{Automatic evaluation results of style reflection metric and response coherency metric using different pre-trained language models. 
}
\label{tab:different_lms}
\vspace{-1em}
\end{table}

%% file: Sections/5_Conclusion.tex
\section{Conclusion}\label{sec:5_conclusion}
In this paper, we introduce the task of mimicking a fictional character by using only a few utterances of the character.
We propose a new method, Pseudo Dialog Prompting, which builds a prompt for a language model to solve this task by creating a pseudo dialog using the given utterance set with a retrieval model.
Extensive experiments show that our method effectively generates responses that reflect the style of a given character better than baseline models and even advantaged models.

%% file: Sections/6_EthicalConsiderations.tex
\section*{Ethical Considerations}\label{sec:6_ethical_consideration}
Like any conversation or generation model, we note that the quality of the models' responses depends on the quality of its training data.
Our Base-LM model was trained on The Pile dataset~\cite{gao2020pile} and Pushshift Reddit dataset~\cite{baumgartner2020pushshift}. Since the contents in these datasets were collected online, they may include underlying biases or potentially offensive words. These biases and toxicities can be projected into our models. Therefore, we highly recommend that additional steps are taken to filter out profanity and inappropriate responses when the model is deployed to the real world.

Furthermore, while we intend our method to be used to mimic fictional characters from movies, shows and stories to build more engaging conversation models, we also recognize that it is possible to use our method to mimic real-life individuals based on their utterances.
Some potential risks include impersonating individuals, which can be harmful to the targeted individuals, and mimicking figures to generate content that can be harmful to groups of individuals.
We hope that our method is deployed in a safe manner to avoid such malicious usage.

%% file: Sections/Appendix.tex
\clearpage
\newpage
\section*{Appendix}

\section{Related Work}
\subsection{Text Style Transfer}
There are various studies of text style transfer, which are not bound for open-domain conversation.
These studies utilize task-specific parallel data for style transfer~\cite{jhamtani2017shakespearizing,rao2018dear,chawla2020semi}.
However, since obtaining parallel data requires a substantial amount of labor, many studies have been proposed to address unsupervised text style transfer recently.

One line of the studies addresses unsupervised text style transfer by constructing pseudo-paired texts and training a model on those paired texts.
\citet{subramanian2018multiple, zhang2018style} create those parallel texts by back-translation and 
\citet{lai2021generic} construct pseudo-parallel paired texts using generic resources and fine-tune two generation models on these pseudo parallel texts iteratively.
However, these methods require a further step to create parallel data by synthesizing or leveraging existing resources and train generation models on those pairs.
Moreover, these methods are not applicable for arbitrary text style transfer since the methods target predefined style pairs only (e.g., British-American and Modern-Shakespeare).

Another line of studies solves unsupervised text style transfer by disentangling content and style from texts.
Most of the studies~\cite{shen2017style,li2018delete} assume that enough style-labeled texts are available for training.
\citet{ma2021collaborative} utilize a collaborative learning framework to disentangle content and style from the texts, but it also requires style-labeled texts while training generation models.
\citet{zhao2018language} consider a scenario where only target style labels are available.
Since our work considers the task where only a few utterances of characters are available to generate responses, we do not consider these methods requiring style-labeled texts as baseline methods of evaluation.
Instead, we select TextSETTR~\cite{riley2021textsettr}, which extracts style vectors from generic texts without requiring style-labeled texts, as a baseline method for a fair evaluation.

\subsection{Stylized Response Generation}
There are several studies that directly address stylized response generation, which is a special case of text style transfer. 
Similar to text style transfer, stylized response generation can also be divided into supervised~\cite{akama2017generating} and unsupervised ways~\cite{gao2019structuring,zheng2020stylized}.
In particular, \citet{gao2019structuring} utilize conversation data with distinct style-labeled texts to models a shared latent space. 
\citet{zheng2020stylized} utilize unpaired texts that have distinct styles and convert them into pseudo conversation pairs using inverse model. 
Finally, these pseudo conversation pairs are employed to train a generation model with a joint training process.
However, the above studies do not meet our problem condition since they require a considerable amount of style-labeled texts or need further training procedure and target only specific styles.
 
Several stylized response generation studies could be applicable to our setting.
\citet{boyd2020large} introduce a method to reflect arbitrary user's style by utilizing the user's conversation history without requiring additional fine-tuning.
\citet{madotto2021few} utilize prompt-based few-shot learning to control style of generated responses.
We extend~\citet{madotto2021few}'s framework to stylized response generation as a baseline method (Zero-shot Prompt) by providing a proper prompt.

\input{Tables/11_PDP_Prompt}
\input{Tables/7_GCC_Prompt}
\section{Model Details}\label{subsec:1_appendix_details}
\textbf{Pseudo Dialog Prompting Details.}
Like all other baseline models, we also employ Base-LM to generate responses by conditioning it with a prompt built by Pseudo Dialog Prompting method.
For the retrieval-based conversation model $R$ used for Pseudo Dialog Prompting, we use a 256M parameter Bi-encoder~\citep{humeau2019poly} retrieval model trained with the method of \citet{kim2021distilling}, along with the utterances of Blended Skill Talk training dataset as the fixed set of context candidates $\mathcal{C}$.
Table~\ref{tab:pdp_prompt} shows the prompt template and an example for the character for Pseudo Dialog Prompting.
\newline
\textbf{Base-LM Training Details.}
The sizes of the datasets are both 700G for the Pile and the Pushshift Reddit comment dataset, respectively.
For the Pushshift Reddit comment dataset, we use the comment created up to April 2020.
For the hyperparameters of the model, we use 32 as the number of layers, 3072 as the number of units in each bottleneck layer, and 32 as the number of attention heads.
For the tokenizer, we use the same byte-level BPE tokenizer as in GPT-2~\citep{radford2019language}.
We use an initial learning rate of $1.6 \times 10^{-4}$ and batch size of 512 for the training hyperparameters and follow other configurations from~\citet{brown2020language}.
The model is trained for a total of 300 billion tokens, which takes approximately 21 days using 64 NVIDIA A100 GPUs.
\newline
\textbf{GCC Training Details.}
We reproduce GCC with three minor modifications: First, we train the model with the HLA-chat dataset instead of the Reddit comment dataset. 
Secondly, we do not include a context (notated 'parent comment' in the original paper) of reference histories since only the utterances of a character are available in our task setup.
Lastly, we do not utilize token-type embeddings since dialogs in HLA-chat only consist of two speakers.
The HLA-Chat dataset is divided into an 8:1:1 split based on character, and each split is used as train, validation, and test split, respectively.
While constructing a dataset, we omit ten characters selected for our evaluation for fair comparison as a baseline.
For reference contexts, we randomly sample a maximum of eight utterances of a character, excluding the gold response itself.
We fine-tune the model from Base-LM using the data format of Table~\ref{tab:gcc_prompt} with the hyperparameter of input length 1024, initial learning rate $1.0\times10^{-5}$ with cosine decay schedule with 100 warmup steps, 10 training epochs, and the batch size 128.
We use the early-stopped model using the validation split perplexity.
\newline
\textbf{Full-dataset Fine-tuning Training Details.}
\input{Tables/5_Full_shot_Prompt}
We fine-tune Base-LM on full HLA-Chat dataset, using a data format of Table~\ref{tab:full_shot_prompt}.
Similar to GCC, HLA-Chat data is divided into an 8:1:1 split, but here ten characters selected for evaluation are contained in the training set.
We fine-tune the model from Base-LM using the hyperparameter of input length 1024, initial learning rate $1.0\times10^{-6}$ with cosine decay schedule with 100 warmup steps, 10 training epochs, and the batch size 128.
We also early-stopped fine-tuning using the validation split perplexity.
\newline
\textbf{Prompts for Baseline Methods.}
\input{Tables/3_NoMatch_Prompt}
\input{Tables/4_ZeroShot_Prompt}
\input{Tables/15_BaseLM_Prompt}
Tables~\ref{tab:no_match_prompt}, \ref{tab:zero_shot_prompt}, \ref{tab:base_lm_prompt} show the prompt template and an example for the character for each baseline methods.
Here, we assume we only have two utterances from the character.

\section{Evaluation Details}\label{sec:appendix_evaluation_details}
\textbf{Decoding Options}
When we generate samples, we adopt a top-k decoding strategy which is widely used for generating diverse and specific responses~\cite{fan2018hierarchical}.
We use $k = 20$ for our top-k sampling.
We choose a minimum beam length and a beam size as 10 and 5, respectively, and use 5-gram beam blocking.
\newline
\textbf{Automatic Evaluation}
For the automatic evaluation, we choose ten characters among all characters included in HLA-Chat.
We construct the test set consisting of 5903 utterances by selecting only utterances with a length of 30 or more from among the utterances from DailyDialog test set.
We use the utterances of the test split of DailyDialog dataset for fixed dialog contexts to construct dialog contexts that are typical and not dependent on specific characters.
For the StyleProb metric, we train a character style classifier using the utterances from ten selected characters in the HLA-chat dataset.
We collect the utterances of ten evaluation characters from the dataset and train a 10-class classifier by fine-tuning the RoBERTa-base model.
We use Huggingface transformers~\citep{wolf-etal-2020-transformers} to train the model, and use the learning rate $2.0 \times 10^{-5}$, batch size 128, the number of training epochs 3.
The accuracy of the classifier on the validation split is 0.5838.
For calculating the MaUdE metric, we use the code officially provided by the authors\footnote{\url{https://github.com/facebookresearch/online_dialog_eval}}.
\newline
\textbf{Human Evaluation}
\input{Figures/Appendix_Figure_1}
For the human evaluation, we select five characters which style can be distinguished apparently.
Additionally, we use the randomly selected subset of the full fixed-context set consisting of 50 contexts.
We use Amazon MTurk for collecting assessments, and Figure~\ref{fig:appendix_1} shows the instructions and the interface for the human evaluation.
We mitigate the bias from the annotator by setting a maximum number of annotations per worker as 20 and randomly shuffling the order of the model and the corresponding response.
To control the annotation quality, we only allow the annotators who satisfy the following requirements: (1) HITs approval rate greater than 95\%, (2) Location is one of Australia, Canada, New Zealand, United Kingdom, and the United States, (3) Lifetime number of HITs approved greater than 1000, following \citet{li2018towards}.
We estimated that each HITs takes around 1.5 minutes on average (87 seconds per each HIT estimated by the 85th percentile of response times) and set the payment to USD 10 per hour. 
Therefore, annotators are paid USD 0.25 per HITs.
\newline
\textbf{Descriptive Statistics.}
\input{Tables/12_DescriptiveStatistics}
We provide the 95\% confidence interval of human evaluation results in Table~\ref{tab:12_descriptive_statistics}.
The 95\% confidence interval of all the MaUdE results reported in the Table~\ref{tab:main_result} is $\pm0.002$.
\newline
\textbf{Dataset Details.}
We mainly used HLA-Chat dataset for our evaluation.
The HLA-Chat dataset is an English single-turn dialogue dataset where the dialogue is scraped from TV show scripts.
Dataset consists of dialogues from 327 characters in 38 TV shows, resulting in a total of 1,042,647 dialogue lines.
We divided the split into 8:1:1 split based on character, where each split is used as train, validation, and test split, respectively.
For our main experiments, we selected ten characters and selected eight utterances that best reveal each character's unique characteristics.
The set of utterances used for describing the characters used for our experiments is reported in our codebase.~\footnote{Attached as supplementary material and will be released open-source afterward.}
\newline
\textbf{Number of Experiments}
We perform the experiment once rather than running it multiple times with different seeds.
Since our evaluation process incorporates a human annotation, which requires a payment to human annotators, we were not able to perform multiple sets of experiments due to the limitation on budget.

\section{Additional Analysis}
\subsection{Lexical Overlap}
\input{Tables/16_Lexical_Overlap}
In Table~\ref{tab:lexical_overlap} we report an additional automated metric, $n$-gram overlap (where $n$ = 2), for analyzing the style of generated responses.
$n$-gram overlap indicates the ratio of n-grams in the generated response, which is contained in the target character utterances.
The trend of $n$-gram overlap metric is similar to that of \textit{StyleProb} metric.
PDP-based methods, especially a Dynamic Match, show higher $n$-gram overlap values than other methods, indicating that PDP-based methods actively utilize the lexical phrases appearing in the character utterances.

The high $n$-gram overlap values of PDP methods indicate that PDP methods actively utilize the lexical phrases appearing in the character utterances.
Using the unique vocabulary of the character will help people to realize a better individualization of the specific character.
Nonetheless, this observation may imply that the model focuses on utilizing lexical language habits and may not capture the inherent characteristics of the character.
Since addressing the inherent characteristics given only a few utterances is a highly challenging task, we think that extending our work to mimic characters' intrinsic characteristics will be an intriguing future direction.

\subsection{More Examples}
\input{Tables/14_Cherry_Picked_With_Baselines}
In Tables~\ref{tab:14_cherry_pick} we show more examples.
We can see that our Static Match and Dynamic Match methods are able to generate responses that contain contents that are highly specific to the character. 
For example, for BMO (from the show Adventure Time) response generated by our method mentions terms such as "core system drivers" and "MO Factory" that are relevant to the fact that BMO is an animated video game console in the show.
Furthermore, we can see that our methods generate a response that reflects the character's style.
For Spock (from Star Trek), our response reflects Spocks' stoic, highly logical, and cold personality.
For Sheldon (from The Big Bang Theory), our response reflects Sheldon's excited speech style.

\section{Failure Modes of Dynamic Match}
\input{Tables/8_Lemon_Picked}
As in we discussed before, there exists a trade-off between the style reflection and response coherency between Static Match and Dynamic Match.
In Tables~\ref{tab:lemon_pick} we show some failure modes of our Dynamic Match method that reveal how Dynamic Match loses the response coherency. 
In the first case, the model generates a response that exhibits a strong character style but is incoherent to the input context. 
In the second case, the model confuses the identity of the speaker so that the model introduces itself as Dr. Leonard Hofstadter. 
Last but not least, when the given input context is highly specific, we see that the generated responses do not reflect the character's style.

\section{Extending to General Style-Controlling Conversation}
\input{Tables/9_StyleEvaluation}
In this section, we extend our methodology to more general style-controlling conversation tasks such as controlling sentiment, emotion, or writing styles, not just mimicking a fictional character.
We test three style-controlling tasks -- controlling sentiment (Positive, Negative), emotion (Anger, Joy), and writing style (Modern, Shakespearean).
For each task, the utterances for defining a style and a style classifier for the evaluation are obtained from the Yelp restaurant review dataset \footnote{Obtained from \url{https://github.com/luofuli/DualRL}}, GoEmotions dataset~\citep{demszky2020goemotions}, and Shakespearean dataset~\citep{xu2012paraphrasing}, respectively.
Style classifier for each task is trained using the same codebase and hyperparameters as in training the character style classifier in the HLA-chat dataset.
We used Style Accuracy rather than StyleProb, following previous literature on style transfer.

The experimental result of general style-controlling conversation tasks is depicted in Table~\ref{tab:9_style_evaluation}.
Similar to mimicking fictional characters, PDP methods show significantly higher style reflection metrics than the baseline methods in general style controlling tasks.
Especially, \textit{Dynamic Match} shows the best style accuracy metric among all the PDP methods, which is also a trend similarly observed in character mimicking experiments.
These results demonstrate that our method is not limited to the character mimicking task but has the ability to be generally applicable to all kinds of style-controlling conversation tasks.
Although the PDP methods have a lower MaUdE score than baseline methods, we believe this tendency is because the MaUdE metric has difficulties evaluating a sentence that strongly reflects a distinctive style, as discussed in the main text.
For instance, reflecting the emotion "Anger" causes the model to generate upper-cased responses (e.g., "I DO NOT WANT TO EAT LUNCH"), which is an out-of-distribution sample when training the MaUdE model.

\section{Multi-turn Chit-chat Examples}
\input{Figures/Appendix_Multiturn_Example_1}
We show some multi-turn conversation examples with the characters generated by our method in Figure~\ref{fig:multiturn_example_1}.

\section{Mimicking a New Character}
\input{Figures/Appendix_Multiturn_Example_Pie}
To show that our method can be generally applied to any fictional characters that do not appear in the pre-training dataset nor the HLA-Chat dataset, we report a conversation example of the PDP method with an imaginary character generated by ourselves.
The character is called \textit{Pie the Duck}, who is a duck character that quacks all the time, likes to eat fish, and enjoys swimming.
We use the following utterances to define the character:
\begin{itemize}
    \setlength\itemsep{0em}
    \item My name is Pie the Duck, Quack Quack!
    \item I really like swimming, Quack! And I am also good at it, Quack!
    \item I like rainy day!! Quack Quack!!
    \item Salmon avocado salad is my favorite food! But... anything made of fish is fine :)
    \item I'm looking at the sky... Will be fishes living in the sky too? Quack.
    \item I'm so cute! Look at my beak!
    \item I'm recently on a diet to better float on water! It's necessary! Quack!
    \item I majored sports, That's why I'm a good swimmer! Quack Quack!
\end{itemize}
Figure~\ref{fig:multiturn_example_pie} shows the example of a multi-turn conversation with Pie the Duck.
As shown in the example, PDP successfully captures the unique style and persona reflected on characters' utterances, including quacking habits, own name, identity as a duck, favorite food, etc., while maintaining a dialog coherency.
\label{sec:appendix}

\section{Scientific Artifacts}
\input{Tables/17_License}

\noindent \textbf{License.}
Table~\ref{tab:17_license} denotes the license of the datasets and pre-trained models that we used for this paper.
Unless for the case where the license is not specified, all of the licenses allow the use of resources for research purposes; therefore, the use of these artifacts in this work is valid.
\newline
\textbf{Intended Use.}
We want to clarify that the \textit{intended use} of pre-trained language models (when specified) is for text generation or fine-tuning to a downstream task; therefore, we are consistent with their intended use.
\newline
\textbf{Description of the Artifacts.}
Blended Skill Talk (BST) dataset is an English open-domain, multi-turn dialogue dataset built to enable conversational agents to use multiple conversational skills (e.g., Using persona information, talk about knowledge, empathetic conversation) in a single conversation.
DailyDialog dataset is an English open-domain, multi-turn dialogue dataset that tries to reflect our daily communication and cover various topics about our daily lives.
We describe the HLA-Chat dataset in Section~\ref{sec:appendix_evaluation_details}.
The Pile dataset is an 800GB text corpus targeted at training large-scale language models, mostly consisting of English texts and constructed from 22 diverse text sources.
The Pushshift Reddit Comment dataset is a dump of comments from the English website Reddit\footnote{\url{https://www.reddit.com}}.
\newline
\textbf{Privacy and Offensive Contents.}
We do not collect any new data that can identify unique people / contain offensive content.
BST, Dailydailog, HLA-Chat dataset is manually created using human annotators or scraped from TV show scripts, therefore having low risk on the issue of privacy or offensive content.
As discussed in their paper, the Pile dataset explicitly used a profanity checker algorithm to reduce the pejorative content.
While processing the Pushshift dataset, we tried to exclude the training offensive contents using blocklist keywords.
Also, we did not include some subreddits that mostly contain offensive content.

%% file: Tables/11_PDP_Prompt.tex
\begin{table*}[t]
\centering
\footnotesize
\begin{tabular}{l}
\toprule
\textbf{Template} \\
\midrule
\multicolumn{1}{l}{\begin{tabular}[l]{l}
\texttt{The below are quotes of \{\{character\_name\}\} during conversation.}\\
\texttt{User: \{\{c1\}\}}\\
\texttt{\{\{character\_name\}\}: \{\{u1\}\}}\\
\texttt{User: \{\{c2\}\}}\\
\texttt{\{\{character\_name\}\}: \{\{u2\}\}}\\
\texttt{User: \{\{x\}\}}\\ 
\texttt{\{\{character\_name\}\}}: 
\end{tabular}} \\
\midrule
\textbf{Example Prompt} \\
\midrule
\multicolumn{1}{l}{\begin{tabular}[l]{l}The below are quotes of Marge Simpson from The Simpsons during conversation.\\User: I think I'm going to give it a try.\\Marge Simpson from The Simpsons: Aw, Homie, you'll always be my western hero.\\User: I'm from Oklahoma so she was a big deal for our state. We've made lots of country music stars.\\Marge Simpson from The Simpsons: Isn't Bart sweet, Homer? He sings like a little angel.\\
User: Okay. what do you want to do? \\Marge Simpson from The Simpsons: \end{tabular}} \\
\bottomrule
\end{tabular}
\caption{Prompt template and example prompt for Pseudo Dialog Prompting.}
\label{tab:pdp_prompt}
\end{table*}

%% file: Tables/7_GCC_Prompt.tex
\begin{table*}[t]
\footnotesize
\centering
\begin{tabular}{l}
\toprule
\textbf{Training Data Template} \\
\midrule
\multicolumn{1}{l}{\begin{tabular}[l]{l}
\texttt{\{\{u1\}\}}\\
\texttt{\{\{u2\}\}}\\
\texttt{\{\{x\}\}<EOT>\underline{\{\{response\}\}<EOT>}}\end{tabular}} \\
\midrule
\textbf{Training Example} \\
\midrule
\multicolumn{1}{l}{\begin{tabular}[l]{l}Aw, Homie, you'll always be my western hero.\\Isn't Bart sweeet, Homer? He sings like a little angel.\\Oh my God! It's like Christmas in December! Let's celebrate now.<EOT>\underline{Homer, please!<EOT>}\end{tabular}} \\
\bottomrule
\end{tabular}
\caption{A template for training data and example for GCC. Model is trained to predict the underlined part given previous context.}
\label{tab:gcc_prompt}
\end{table*}

%% file: Tables/5_Full_shot_Prompt.tex
\begin{table*}[t]
\footnotesize
\centering
\begin{tabular}{l}
\toprule
\textbf{Training Data Template} \\
\midrule
\multicolumn{1}{l}{\begin{tabular}[l]{l}
\texttt{\{\{character\_name\}\}}\\ 
\texttt{\{\{x\}\}<EOT>\underline{\{\{response\}\}<EOT>}}\end{tabular}} \\
\midrule
\textbf{Training Example} \\
\midrule
\multicolumn{1}{l}{\begin{tabular}[l]{l}Marge Simpson from The Simpsons\\Oh my God! It's like Christmas in December! Let's celebrate now.<EOT>\underline{Homer, please!<EOT>}\end{tabular}} \\
\bottomrule
\end{tabular}
\caption{A template for training data and example for Full-dataset Fine-tuning. Model is trained to predict the underlined part given previous context.}
\label{tab:full_shot_prompt}
\end{table*}

%% file: Tables/3_NoMatch_Prompt.tex
\begin{table*}[t]
\centering
\footnotesize
\begin{tabular}{l}
\toprule
\textbf{Template} \\
\midrule
\multicolumn{1}{l}{\begin{tabular}[l]{l}
\texttt{The below are quotes of \{\{character\_name\}\} during conversation.}\\
\texttt{- \{\{u1\}\}}\\
\texttt{- \{\{u2\}\}}\\
\texttt{The below are conversation between User and \{\{character\_name\}\}.}\\
\texttt{User: \{\{x\}\}}\\ 
\texttt{\{\{character\_name\}\}}: 
\end{tabular}} \\
\midrule
\textbf{Example Prompt} \\
\midrule
\multicolumn{1}{l}{\begin{tabular}[l]{l}The below are quotes of Marge Simpson from The Simpsons during conversation.\\$-$ Aw, Homie, you'll always be my western hero.\\$-$ Isn't Bart sweet, Homer? He sings like a little angel.\\The below are conversation between User and Marge Simpson from The Simpsons.\\
User: Okay. what do you want to do? \\Marge Simpson from The Simpsons: \end{tabular}} \\
\bottomrule
\end{tabular}
\caption{Prompt template and example prompt for Only Utterances.}
\label{tab:no_match_prompt}
\end{table*}

%% file: Tables/4_ZeroShot_Prompt.tex
\begin{table*}[t]
\centering
\footnotesize
\begin{tabular}{l}
\toprule
\textbf{Template} \\
\midrule
\multicolumn{1}{l}{\begin{tabular}[l]{l}
\texttt{Dialogue:}\\
\texttt{User: \{\{x\}\}}\\
\texttt{\{\{character\_name\}\}:}\end{tabular}} \\
\midrule
\textbf{Example Prompt} \\
\midrule
\multicolumn{1}{l}{\begin{tabular}[l]{l}Dialogue:\\User: Okay. what do you want to do? \\Marge Simpson from The Simpsons: \end{tabular}} \\
\bottomrule
\end{tabular}
\caption{Prompt template and example prompt for Zero-shot Prompt.}
\label{tab:zero_shot_prompt}
\end{table*}

%% file: Tables/15_BaseLM_Prompt.tex
\begin{table*}[t]
\centering
\footnotesize
\begin{tabular}{l}
\toprule
\textbf{Template} \\
\midrule
\multicolumn{1}{l}{\begin{tabular}[l]{l}
\texttt{Dialogue:}\\
\texttt{User: \{\{x\}\}}\\
\texttt{Guest:} \end{tabular}} \\
\midrule
\textbf{Example Prompt} \\
\midrule
\multicolumn{1}{l}{\begin{tabular}[l]{l}Dialogue:\\User: Okay. what do you want to do? \\Guest: \end{tabular}} \\
\bottomrule
\end{tabular}
\caption{Prompt template and example prompt for Base-LM when used to generate responses for TextSETTR method.}
\label{tab:base_lm_prompt}
\end{table*}

%% file: Figures/Appendix_Figure_1.tex
\begin{figure*}[t]
\centering
\includegraphics[width=\textwidth]{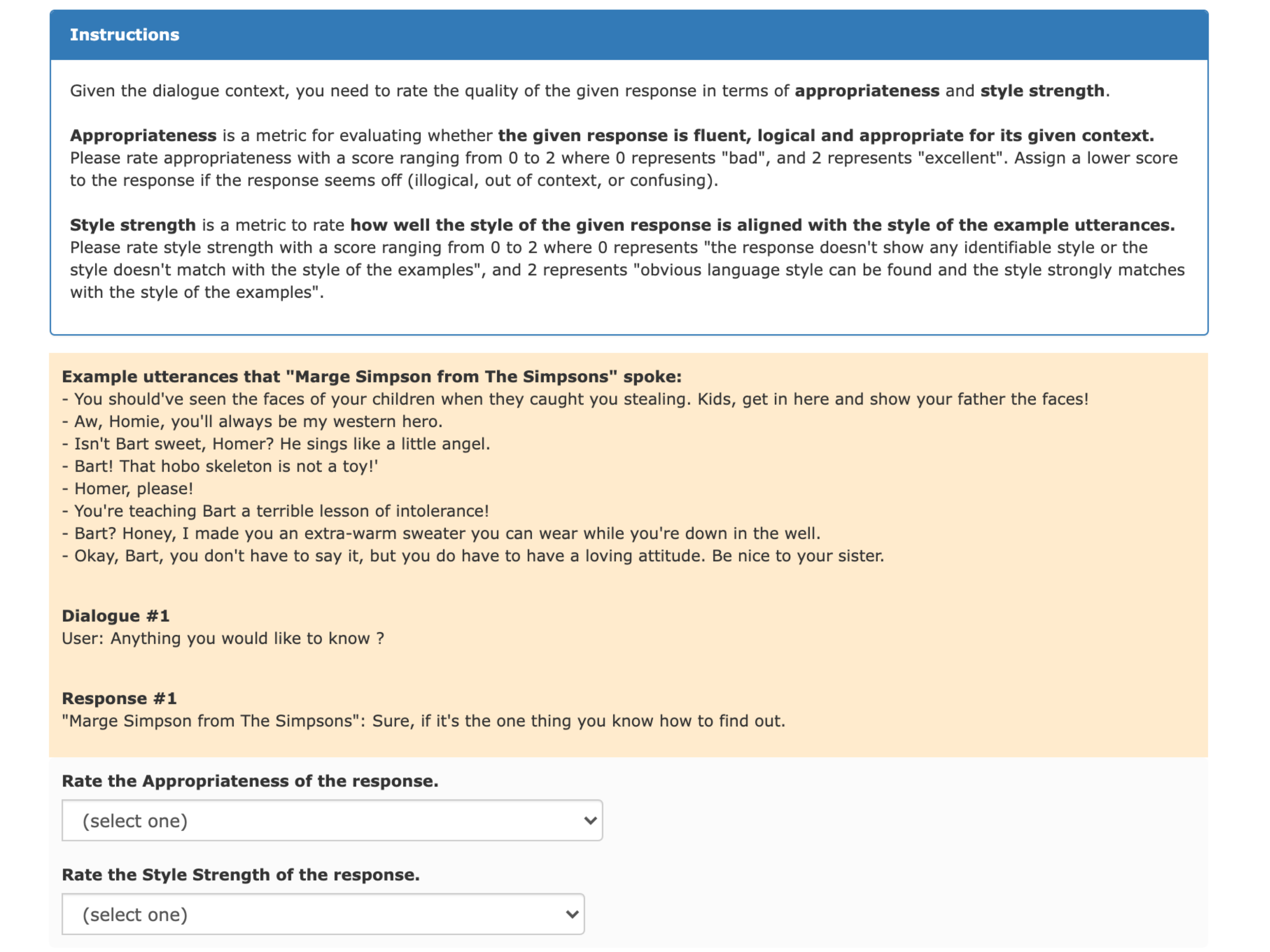}
\caption{The interface of human evaluation for appropriateness and style strength.}
\label{fig:appendix_1}
\end{figure*}

%% file: Tables/12_DescriptiveStatistics.tex
\begin{table*}[t]\setlength{\tabcolsep}{0.4em}
\centering
\footnotesize
\begin{tabular}{clcccccccc}
\toprule
\multicolumn{1}{c}{\multirow{2}[2]{*}{Method Type}} & \multicolumn{1}{c}{\multirow{2}[2]{*}{Methods}} & \multicolumn{2}{c}{Human Evaluation (Raw)} & \multicolumn{2}{c}{Human Evaluation (Cali.)} &  \\ \cmidrule(lr){3-4}  \cmidrule(lr){5-6}
& & \multicolumn{1}{c}{Style.} & \multicolumn{1}{c}{Appr.}  & \multicolumn{1}{c}{Style.} & \multicolumn{1}{c}{Appr.} \\
\midrule
\multirow{4}{*}{\textbf{Baselines}}
& Only Utterances& 1.200$\pm$0.052 & 1.263$\pm$0.049 & 1.147$\pm$0.013 & 1.124$\pm$0.013\\
& Zero-shot Prompt& 1.172$\pm$0.051& 1.236$\pm$0.048 & 1.114$\pm$0.012 & 1.037$\pm$0.014\\
& TextSETTR& 1.155$\pm$0.051& 1.317$\pm$0.050 & 1.060$\pm$0.014 & 1.059$\pm$0.013 \\
& GCC& 1.193$\pm$0.051 & 1.256$\pm$0.048 & 1.136$\pm$0.013 & 1.069$\pm$0.014 \\
\midrule
\multirow{3}{*}{\textbf{Ours}}
& Static Match & 1.277$\pm$0.052& 1.229$\pm$0.052 & 1.195$\pm$0.013 & 1.123$\pm$0.014\\
& Dynamic Match& 1.331$\pm$0.049& 1.203$\pm$0.051& 1.276$\pm$0.013 & 1.021$\pm$0.013 \\
& Random Match& 1.297$\pm$0.050 & 1.195$\pm$0.053 & 1.199$\pm$0.013 & 1.035$\pm$0.014 \\
\midrule
\midrule
\multirow{2}{*}{\textbf{Advantaged}}
& Full-dataset Fine-tuning& 1.260$\pm$0.051 & 1.329$\pm$0.048 & 1.164$\pm$0.013 & 1.130$\pm$0.013\\
& Gold Match &1.287$\pm$0.050 & 1.245$\pm$0.051 & 1.157$\pm$0.012 & 1.097$\pm$0.013 \\
\bottomrule
\end{tabular}
\caption{Evaluation results of Human evaluation results with 95\% confidence interval.
Raw and Cali. indicate raw human evaluation results and Bayesian calibrated scores, respectively.
}
\vspace*{-4mm}
\label{tab:12_descriptive_statistics}
\end{table*}

%% file: Tables/16_Lexical_Overlap.tex
\begin{table}
\centering
\footnotesize
\begin{tabular}{clc}
\toprule
Method Type & Methods & $n$-gram overlap \\
\midrule
\multirow{4}{*}{\textbf{Baselines}}
& Only Utterances& 0.0417\\
& Zero-shot Prompt& 0.0368\\
& TextSETTR& 0.0222 \\
& GCC& 0.0632 \\
\midrule
\multirow{3}{*}{\textbf{Ours}}
& Static Match & 0.1856 \\
& Dynamic Match& 0.3478 \\
& Random Match& 0.1353 \\
\midrule
\midrule
\multirow{2}{*}{\textbf{Advantaged}}
& Full-dataset Fine-tuning& 0.0951\\
& Gold Match & 0.2631 \\
\bottomrule
\end{tabular}
\caption{Evaluation results of $n$-gram overlap between generated response and character utterances.
}
\vspace*{-2mm}
\label{tab:lexical_overlap}
\end{table}

%% file: Tables/14_Cherry_Picked_With_Baselines.tex
\begin{table*}[h]
\centering
\footnotesize
\setlength
\tabcolsep{2.5pt}
\begin{tabular}{{p{4.5cm} l p{8cm}}}
\toprule
\textbf{Input Context} & \textbf{Method} & \textbf{Character \& Response} \\
\cmidrule{1-3}
\multirow{8}{*}{All you have to do is ask for help} & \textbf{Static Match (Ours)} & \textbf{BMO}: I am not able to ask anyone help. It's a little more complicated than that. \\
\cmidrule(lr){2-3}
& \textbf{Dynamic Match (Ours)}& \textbf{BMO}: Okay. Please take me to get fixed. I need to get new core system drivers installed. We can get them at the MO Factory in the Bad Lands, where I was born \\
\cmidrule[\heavyrulewidth]{2-3}
 & Only Utterance & \textbf{BMO}: If your friend was in trouble, he would know what to do.  \\
\cmidrule(lr){2-3}
 & Zero-shot Prompt & \textbf{BMO}: I don't think I'm ready. I don't think anyone is ready.\\
\cmidrule(lr){2-3}
 & GCC & \textbf{BMO}: All I have to do is ask for your help.\\
\cmidrule(lr){2-3}
 & Base-LM + TextSETTR & \textbf{BMO}: When I was in this situation, I asked for help from a friend.\\
\cmidrule[\heavyrulewidth]{2-3}
 & Full-dataset Fine Tuning & \textbf{BMO}: Oh, okay. But how do you know it's the right question? \\
\cmidrule(lr){2-3}
 & Gold Match & \textbf{BMO}: Okay, but you got to go first. Please! \\
\cmidrule[\heavyrulewidth]{1-3}
\multirow{8}{*}{\makecell[l]{Did you know that he has \\20 years teaching experience?}} & \textbf{Static Match (Ours)} & \textbf{Spock}: You're right, Mister Scott. He knows every medical procedure there is. Spock out. \\
\cmidrule(lr){2-3}
&\textbf{Dynamic Match (Ours)}& \textbf{Spock}: I have already begun investigation into that possibility. \\
\cmidrule[\heavyrulewidth]{2-3}
 & Only Utterance & \textbf{Spock}: Yes, but I never met him. Do you know where he lives?  \\
\cmidrule(lr){2-3}
 & Zero-shot Prompt & \textbf{Spock}: The Next Generation: No. I didn't. Who's he?\\
\cmidrule(lr){2-3}
 & GCC & \textbf{Spock}: Yes, sir. I have heard many stories about Spock.\\
\cmidrule(lr){2-3}
 & Base-LM + TextSETTR & \textbf{Spock}: It's not easy to do that.\\
\cmidrule[\heavyrulewidth]{2-3}
 & Full-dataset Fine Tuning & \textbf{Spock}: Yes, and he has an encyclopedic knowledge of the Federation. \\
\cmidrule(lr){2-3}
 & Gold Match & \textbf{Spock}: I did not. How did you arrive at that number? \\
 \cmidrule[\heavyrulewidth]{1-3}
 \multirow{8}{*}{\makecell[l]{I would prefer Chinese food \\for this party.}} & \textbf{Static Match (Ours)} & \textbf{Sheldon}: Well, we've got chicken, but I'm not sure if it's kosher. \\
 \cmidrule(lr){2-3}
 & \textbf{Dynamic Match (Ours)}& \textbf{Sheldon}: Well, I've spent the last three hours in an online debate in the DC Comics Batman chatroom, and I need you, but I guess Chinese food would be good. \\
\cmidrule[\heavyrulewidth]{2-3}
 & Only Utterance & \textbf{Sheldon}: Well, there are a couple of Chinese food options, but they're both kind of pricey. \\
\cmidrule(lr){2-3}
 & Zero-shot Prompt & \textbf{Sheldon}: I don't think I'm ready. I don't think anyone is ready.\\
\cmidrule(lr){2-3}
 & GCC & \textbf{Sheldon}: Oh? Why? What happened? What's the matter?\\
\cmidrule(lr){2-3}
 & Base-LM + TextSETTR & \textbf{Sheldon}: But this time, it's just the two of us.\\
\cmidrule[\heavyrulewidth]{2-3}
 & Full-dataset Fine Tuning & \textbf{Sheldon}: All right. What's the difference, right? \\
\cmidrule(lr){2-3}
 & Gold Match & \textbf{Sheldon}: I am so sorry. I am so so sorry.\\
 \bottomrule
\end{tabular}
\caption{Comparison between responses generated using our Static Match, Dynamic Match, and baseline methods. BMO is from Adventure Time, Spock is from Star Trek, and Sheldon is from The Big Bang Theory. }
\label{tab:14_cherry_pick}
\end{table*}

%% file: Tables/8_Lemon_Picked.tex
\begin{table*}[h]
\centering
\footnotesize
\setlength
\tabcolsep{2.5pt}
\begin{tabular}{{p{4.5cm} p{8cm}}}
\toprule
\textbf{Input Context} & \textbf{Character \& Response} \\
\cmidrule[\heavyrulewidth]{1-2}
Oh , I get it . When did we decide to meet? & \textbf{Spock} (from \textit{Star Trek}): I'll scan the ship for more information. Spock out. \\
\cmidrule{1-2}
Our neighbor, Bruce, will take part in the relay race. Let's go have a look. & \textbf{Sheldon} (from \textit{The Bigbang Theory}): Hey everyone. I'm Dr. Leonard Hofstadter. And today I'd like to share with you my theories about the nature of quantum mechanics. \\
\cmidrule{1-2}
And the total ratio of return on assets should not be lower than the average, for that particular industry, for the first two quarters of the year. & \textbf{Marge Simpon} (from \textit{The Simpsons}): You need to stop making these ridiculous arguments. You don't know what you're talking about. \\
\bottomrule
\end{tabular}
\caption{Failure modes for responses generated by using Dynamic Match for each character.}
\label{tab:lemon_pick}
\end{table*}

%% file: Tables/9_StyleEvaluation.tex
\begin{table*}[t]
\centering
\footnotesize
\begin{tabular}{clccc}
\toprule
\multicolumn{1}{c}{\multirow{1}{*}{Task}} & \multicolumn{1}{c}{\multirow{1}{*}{Method}} & \multicolumn{1}{c}{Style Accuracy(\%)} & \multicolumn{1}{c}{MaUdE}\\
\midrule
\multirow{5}{*}{\shortstack[c]{\textbf{Sentiment}\\\textbf{(Positive, Negative)}}}
& Only Utterances & 67.34 & 0.9033\\
& Zero-shot Prompting & 60.54 & 0.9089\\
\cmidrule(lr){2-4}
& Static Match & 71.54 & 0.8668\\
& Dynamic Match & 79.37 & 0.8356\\
& Random Match & 77.87 & 0.8141\\
\midrule
\multirow{5}{*}{\shortstack[c]{\textbf{Emotion}\\\textbf{(Anger, Joy)}}}
& Only Utterances & 79.54 & 0.8139\\
& Zero-shot Prompting & 71.08 & 0.8965 \\
\cmidrule(lr){2-4}
& Static Match & 86.22 & 0.6413\\
& Dynamic Match & 89.99 & 0.6220\\
& Random Match & 88.28 & 0.6412\\
\midrule
\multirow{5}{*}{\shortstack[c]{\textbf{Writing Style}\\\textbf{(Modern, Shakespearean)}}}
& Only Utterances & 53.63 & 0.9055\\
& Zero-shot Prompting & 51.07 & 0.8946\\
\cmidrule(lr){2-4}
& Static Match & 88.75 & 0.8368\\
& Dynamic Match & 92.95 & 0.7926\\
& Random Match & 90.48 & 0.8204 \\
\bottomrule
\end{tabular}
\caption{Automatic evaluation results of style reflection metric and response coherency metric on different style controlling tasks. 
\textit{Gold Match} is omitted since the datasets do not contain the ground-truth context.}
\label{tab:9_style_evaluation}
\end{table*}

%% file: Figures/Appendix_Multiturn_Example_1.tex
\begin{figure*}[t]
\centering
\includegraphics[width=0.7\linewidth]{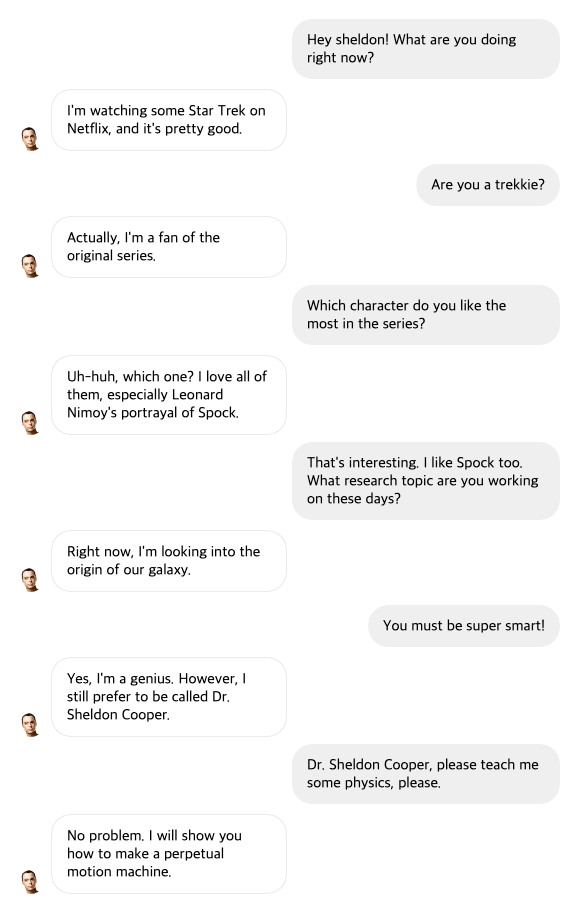}
\caption{Cherry-picked example of multi-turn conversation with the Sheldon Cooper from \textit{the Bigbang Theory} (left speaker) and the paper author (right speaker), generated by the Static Match method.}
\label{fig:multiturn_example_1}
\end{figure*}

%% file: Figures/Appendix_Multiturn_Example_Pie.tex
\begin{figure*}[t]
\centering
\includegraphics[width=0.7\linewidth]{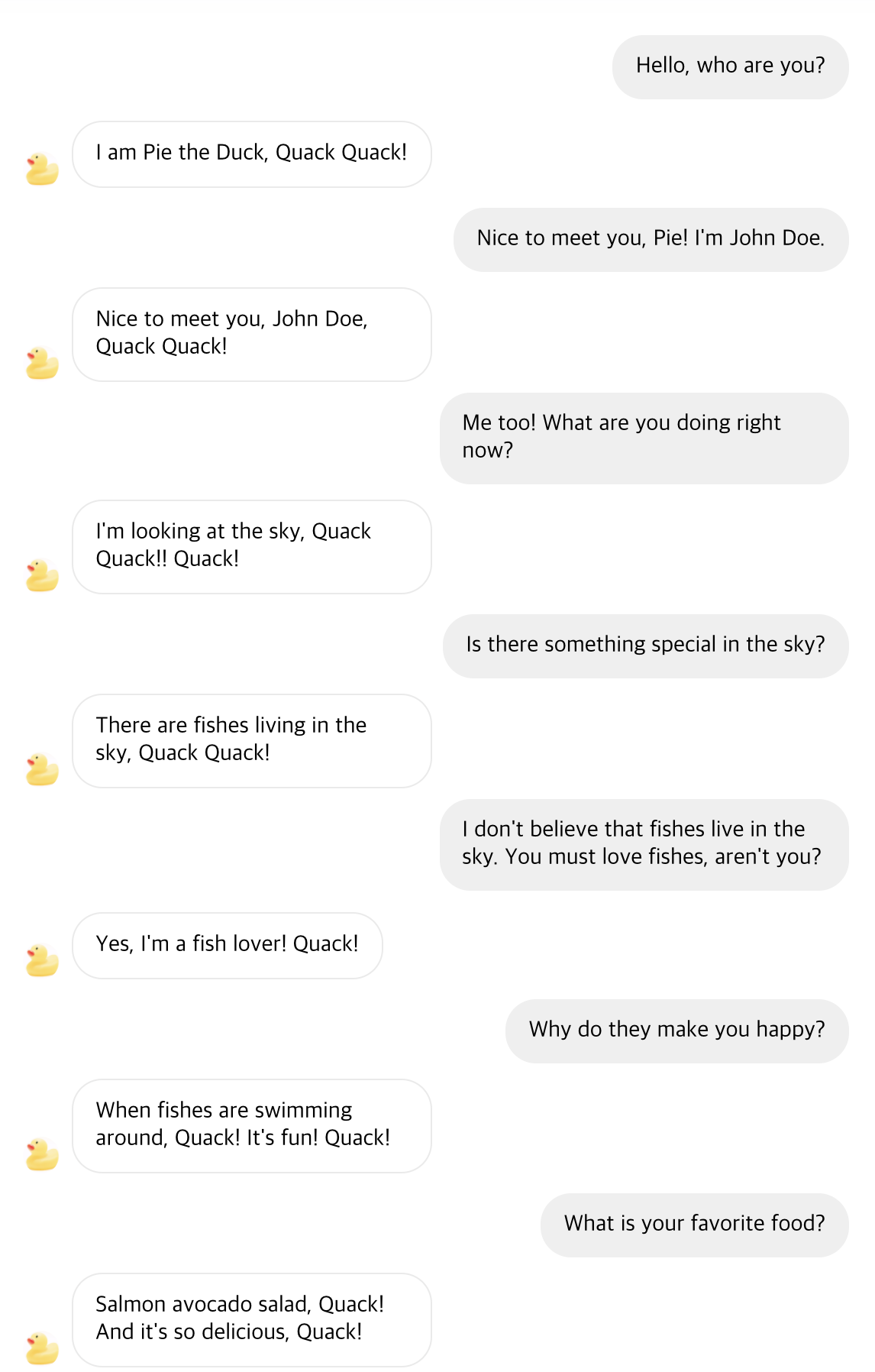}
\caption{Cherry-picked example of multi-turn conversation with the imaginary character \textit{Pie the Duck} and the paper author (right speaker), generated by the Dynamic Match method.}
\label{fig:multiturn_example_pie}
\end{figure*}

%% file: Tables/17_License.tex
\begin{table*}[t]
\centering
\footnotesize
\begin{tabular}{clcc}
\toprule
\multicolumn{1}{c}{\multirow{1}{*}{Artifact Type}} & \multicolumn{1}{c}{\multirow{1}{*}{Name}} & \multicolumn{1}{c}{License} & \multicolumn{1}{c}{Approves the use for research}\\
\midrule
\multirow{5}{*}{\shortstack[c]{\textbf{Dataset}}}
& Blended Skill Talk & CC-BY-4.0 & O \\
& HLA-Chat & Not specified & ? \\
& The Pile & MIT & O \\
& Pushshift Reddit & Not specified & ? \\
& DailyDialog & CC-BY-NC-SA 4.0 & O \\
\midrule
\multirow{3}{*}{\shortstack[c]{\textbf{Pre-trained LM weights}}}
& GPT-J & Apache 2.0 & O \\
& GPT-Neo & Apache 2.0 & O \\
& GPT2-xl & MIT & O\\
\bottomrule
\end{tabular}
\caption{License of the scientific artifacts that we used in this paper.}
\label{tab:17_license}
\end{table*}